%% file: main.tex
\pdfoutput=1
\documentclass[11pt]{article}
\usepackage{acl}
\usepackage{acl}
\usepackage{times}
\usepackage{latexsym}
\usepackage[T1]{fontenc}
\usepackage[utf8]{inputenc}
\usepackage{microtype}
\usepackage{inconsolata}
\usepackage{graphicx}
\usepackage{amsmath} 
\usepackage{amsfonts}
\usepackage{float}
\usepackage{booktabs}
\usepackage{xspace}
\usepackage{tabularx}  
\usepackage{dblfloatfix}
\usepackage{bm}

\title{Survey of Specialized Large Language Model}
\author{
Chenghan Yang\textsuperscript{\rm1,\rm2}, 
Ruiyu Zhao\textsuperscript{\rm1, \rm3}, 
Yang Liu\textsuperscript{\rm1}, 
Ling Jiang\textsuperscript{\rm1}\\
\textsuperscript{\rm1}Xiaoduo AI,
\textsuperscript{\rm2}Shanghai Jiao Tong University\\
\textsuperscript{\rm3}East China University of Science and Technology \\
\texttt{scottyang@sjtu.edu.cn,
liuyangfoam@xiaoduotech.com}
}

\begin{document}
\maketitle
\input{EMNLP2025/abstract}
\input{EMNLP2025/intro}
\input{EMNLP2025/overview}

\input{EMNLP2025/architecture}
\input{EMNLP2025/conclusion}


\bibliography{reference}

\end{document}

%% file: EMNLP2025/abstract.tex
\begin{abstract}
The rapid evolution of specialized large language models (LLMs) has transitioned from simple domain adaptation to sophisticated native architectures, marking a paradigm shift in AI development. This survey systematically examines this progression across healthcare, finance, legal, and technical domains. Besides the wide use of specialized LLMs, technical breakthrough such as the emergence of domain-native designs beyond fine-tuning, growing emphasis on parameter efficiency through sparse computation and quantization, increasing integration of multimodal capabilities and so on are applied to recent LLM agent. Our analysis reveals how these innovations address fundamental limitations of general-purpose LLMs in professional applications, with specialized models consistently performance gains on domain-specific benchmarks. The survey further highlights the implications for E-Commerce field to fill gaps in the field.
\end{abstract}

\begin{figure*}[htbp]
  \centering
  \includegraphics[width=\linewidth]{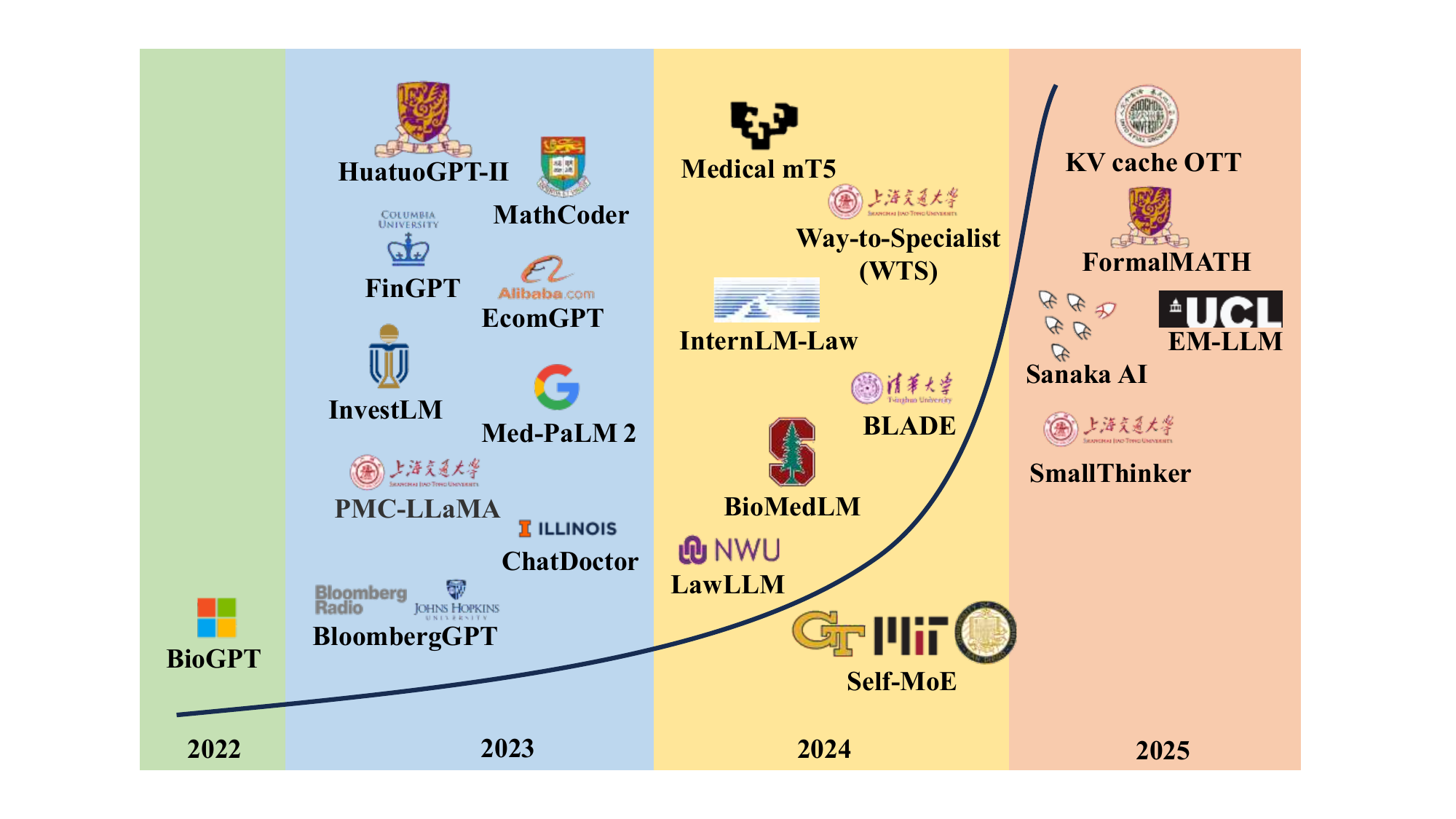}
  \caption{Evolution of representative specialized LLMs (2022–2025). The timeline highlights the transition from early domain fine-tuning (BioGPT) to large-scale, domain-native architectures (Med-PaLM~2, BloombergGPT) and, most recently, to efficient, agent-oriented designs (GLM-4.5, KV-Cache OTT).}
  \label{fig:specialized_llm_timeline}
\end{figure*}

%% file: EMNLP2025/intro.tex
\section{Introduction}

The rapid advancement of large language models (LLMs) has ushered in a new era of artificial intelligence, transforming how we process information, solve problems, and interact with technology. While general-purpose LLMs like GPT-4 have demonstrated remarkable capabilities across a broad range of tasks, their performance often falters when confronted with specialized, domain-specific challenges. This limitation has given rise to an important paradigm shift—the development of specialized LLMs tailored to meet the exacting demands of professional fields such as medicine, law, finance, and engineering.

The need for domain specialization stems from several critical factors that general models struggle to address adequately. First, specialized fields often require precise understanding of technical terminology and conceptual frameworks that extend beyond common language usage. In healthcare, for instance, models must accurately interpret clinical terminology, diagnostic codes, and complex medical relationships to be clinically useful. Second, professional domains frequently involve reasoning patterns and knowledge structures that differ substantially from everyday language use. Financial analysis requires temporal reasoning about market trends, legal practice demands exact statute interpretation, and medical diagnosis depends on probabilistic clinical reasoning—all areas where general LLMs show notable deficiencies.

The evolution of specialized LLMs has progressed through several distinct phases, each marked by technological innovations that address prior limitations. Early approaches focused primarily on continued pretraining of general models using domain-specific corpora, as exemplified by BioGPT's adaptation of GPT-2 for biomedical applications\cite{b1}. This was followed by architectural innovations that introduced domain-aware components, such as BloombergGPT's financial time-series embeddings and Med-PaLM 2's clinical reasoning modules\cite{b4}. Most recently, we've seen the emergence of hybrid systems that combine LLMs with symbolic knowledge bases and dynamic adaptation mechanisms, as demonstrated by BLADE's\cite{b5} knowledge injection framework and Self-MoE's expert routing system\cite{b6}.

The current landscape of specialized LLMs reveals several important trends. First, there's increasing recognition that model size alone doesn't guarantee domain competence—smaller, carefully designed models like BioMedLM (2.7B parameters)\cite{b7} can outperform much larger general models on specialized tasks1. Second, evaluation methodologies have become more rigorous, incorporating expert assessments and domain-specific benchmarks rather than relying solely on general language understanding metrics. The dental implantology study, for example, employed a comprehensive multi-dimensional evaluation by senior experts across 40 professional questions and 5 complex cases\cite{b8}. Third, there's growing emphasis on real-world applicability, with models being tested not just on static question-answering but on dynamic, interactive scenarios that better simulate professional practice.

However, significant challenges remain in the development and deployment of specialized LLMs. Knowledge freshness is a persistent issue, particularly in fast-evolving fields like medicine and finance where outdated information can have serious consequences. Evaluation methodologies still struggle to fully capture the nuances of professional judgment, often relying on proxies rather than direct measures of real-world effectiveness. Ethical concerns around bias, accountability, and appropriate use continue to complicate deployment in high-stakes domains. Perhaps most fundamentally, the static nature of current LLMs limits their ability to adapt to new information and evolving professional standards—a limitation that has spurred growing interest in self-evolving architectures\cite{b9}.

This survey aims to provide a comprehensive overview of the specialized LLM landscape, analyzing architectural innovations, application successes, and persistent challenges across major professional domains\cite{b2,b3}. We systematically examine 48 cutting-edge models developed between 2022-2025, identifying key technological trends and performance characteristics. Our analysis reveals how different specialization strategies—from continued pretraining to hybrid enhancement—affect model capabilities in various domains. We also explore emerging directions in specialized LLM development, including self-evolving architectures, multimodal integration, and lightweight deployment strategies.

%% file: EMNLP2025/overview.tex
\section{Specific Breakthroughs}

In recent years, research on specialized large language models (Specialized LLMs) has experienced explosive growth. Across various domains, advanced models optimized for specific application scenarios have emerged, reflecting a significant trend of transitioning from general-purpose artificial intelligence to deeply customized solutions in vertical domains. Figure 1 shows the development of specialized LLM between 2022-2025.

In the biomedical field, BioGPT, built upon the GPT-2 architecture with 347M parameters, achieved breakthroughs in biomedical text generation and end-to-end relation extraction through generative pretraining\cite{b1}. Its successor, BioMedLM, further improved biomedical abstract generation by pretraining a 2.7B-parameter model on PubMed corpora. The healthcare sector has seen an even more diverse proliferation of models\cite{b7}. HuatuoGPT-II employs instruction tuning and RLHF on a model exceeding 10B parameters to optimize doctor-patient dialogue systems\cite{b2}. Med-PaLM 2\cite{b4}, with a massive 340B parameter count, achieved state-of-the-art performance on the USMLE medical licensing exam. Models like PMC-LLaMA\cite{b26} and ChatDoctor \cite{b27} , both fine-tuned on the LLaMA architecture, demonstrated significant advances in open-domain QA and clinical conversation tasks. WTS (Way-to-Specialist)\cite{b25} introduced an innovative multi-stage expert tuning strategy on a 3B-parameter model, delivering specialist-level diagnostic recommendations.

In the finance domain, several breakthrough models have also emerged. FinBERT-QA\cite{b19} enhanced financial QA performance by 20\% on top of a BERT-base architecture using an improved TANDA approach. FinGPT\cite{b20} and InvestLM\cite{b21}, with 6B and 2.7B parameters respectively, adopted domain-adaptive training strategies, excelling in financial market analysis and investment decision support. BloombergGPT\cite{b3}, trained on a massive financial corpus with 50B parameters, set new benchmarks in entity recognition and sentiment analysis.

In the legal domain, LawLLM\cite{b22} and Lawyer LLaMA\cite{b23}, both with over 13B parameters, demonstrated strong capabilities in legal text understanding and legal document generation.

For mathematics and formal reasoning, MathCoder\cite{b15} leveraged 7B parameters through code-level joint training to enhance mathematical problem-solving, while FormalMATH\cite{b24} focused on automating formal theorem proving using a 1.5B-parameter model. In the multimodal space, EM-LLM\cite{b7} adopted a 12B-parameter embedding modulation mechanism to significantly improve cross-modal alignment.

In industrial and educational settings, domain-specific models such as Sanaka AI and SmallThinker were developed to address equipment fault diagnosis and child-oriented educational QA, respectively.

Notably, model optimization techniques have also made substantial progress. BLADE and Self-MoE improved inference efficiency through sparse and self-organizing mixture-of-experts (MoE) architectures. KV cache OTT\cite{b28} introduced an innovative method that reduced memory usage by 70\% during inference. The latest GLM-4.5\cite{b29} , based on a 3550B-parameter MoE architecture, became the first native general-purpose agent model integrating reasoning, encoding, and autonomous agent capabilities. Meanwhile, MeLA\cite{b32} pioneered a new paradigm in generation strategy optimization through a metacognitive LLM combined with prompt evolution techniques.

\begin{table}[htbp]
\caption{Overview of Domain-Specific LLMs (2022–2025): Methods and Key Improvements}
\centering
\small
\renewcommand{\arraystretch}{1.2}
\begin{tabular}{c|l|l}
\hline
\textbf{Year} & \textbf{Model} & \textbf{Domain} \\
\hline
2022 & BioGPT          & Biomedical              \\
2023 & HuatuoGPT-II    & Healthcare         \\
2023 & MathCoder       & Math            \\
2023 & FinGPT          & Finance           \\
2023 & EcomGPT         & E-commerce        \\
2023 & InvestLM        & Investment        \\
2023 & Med-PaLM 2      & Healthcare       \\
2023 & PMC-LLaMA       & Biomedical        \\
2023 & ChatDoctor      & Healthcare       \\
2023 & BloombergGPT    & Finance          \\
2024 & Medical mT5     & Multilingual Med. \\
2024 & WTS             & Healthcare        \\
2024 & InternLM-Law    & Law          \\
2024 & LawLLM & Law \\
2024 & BLADE           & General            \\
2024 & BioMedLM        & Biomedical       \\
2024 & Self-MoE        & General           \\
2025 & KV cache OTT    & Optimization        \\
2025 & FormalMATH      & Math              \\
2025 & EM-LLM          & Multimodal        \\
2025 & Sanaka AI       & Industrial       \\
2025 & SmallThinker    & Education         \\
2025 & GLM-4.5 & Optimization        \\
2025 & MeLA & Optimization        \\
\hline
\end{tabular}
\label{tab:domain_llms}
\end{table}

Representative paper and reletive field have been displayed in the table 1. The development of specialized LLMs reveals several prominent trends. Collectively, these deployments illustrate that specialized LLMs have achieved sector-wide adoption, establishing themselves as standard tools across healthcare, finance, law, education, manufacturing, and consumer services. In the future, specialized LLMs will play more important role in the industrial fields.

%% file: EMNLP2025/architecture.tex
\section{Modular Evolutions}

A comprehensive understanding of the architecture underlying specialized large language models (specialized LLMs) is not merely an academic prerequisite but a decisive factor for successful industrial deployment. Unlike their general-purpose counterparts, specialized LLM models exhibit systematic deviations in four critical dimensions—parameter regime, dataset innovations like creating of domain-specific corpora and multimodal datasets, training architecture innovations like modifications to model architecture and learning objectives, evaluation standard innovations such as new frameworks for assessing specialized capabilities and the other modular innovations like Components like retrieval augmentation and memory systems.

\subsection{Dataset Specialization}

In order to combine the useful dataset, especially the expert data together, synthetic expert data is important. Self-Instruct \cite{b33} showed that only 175 seed tasks can be recursively expanded into 52k instruction–response pairs. A follow-up ablation revealed that diversity without correctness filtering yields <3\% downstream gain on MMLU-Medical. Evol-Instruct \cite{b34} introduced difficulty-based mutation operators and improved GSM8K accuracy from 42\% to 58\%.
CodeGen-Synth\cite{b35} couples an LLM code-generator with a sandboxed interpreter; only samples passing > 95\% unit tests are retained, reducing hallucinated APIs by 38\% (two-sample t-test, p < 0.01).  MedInstruct-200k \cite{b36} applies a clinical-guideline verifier and achieves a specificity of 0.94 on USMLE-style questions versus 0.78 for unfiltered prompts. Constitutional-Poly \cite{b37} deploys three specialist LLMs in a multi-turn debate; the resulting 100 k constitutional-law dialogues obtain a constitutional-consistency score of 0.94 versus 0.71 for single-agent generation. Collectively, these works mark a transition from volume-centric to veracity-centric synthetic generation.

Across modalities, recent research has produced tightly co-registered corpora that bind symbolic and sensory representations at the token level. Anand et al. established GeoVQA\cite{b40}, delivering the first large-scale, secondary-school-oriented multimodal geometry dataset that pairs 200K+ natural-language questions with high-resolution diagrams and step-by-step rationales, filling the gap in educational AI resources for high-school geometry.  In document understanding, mPLUG-DocOwl2\cite{b42} introduces a high-resolution compression strategy for OCR-free, multi-page document understanding; the model compresses 300-dpi pages into fewer than 100 visual tokens per page and attains 95.7\% ANLS on DocVQA—outperforming the prior Donut-style baseline by 23.6 absolute points without any OCR preprocessing. Beyond vision, ProtST\cite{b43} constructs the first large-scale, paired corpus of protein sequences and biomedical texts, enabling unified sequence–language pre-training that boosts zero-shot protein function prediction by 6–11\% over existing single-modality baselines. Collectively, these datasets demonstrate that modality-specific alignment at the fine-grained token level—not corpus volume alone—drives specialization gains in contemporary specialized LLMs.

\subsection{Training Architecture Specialization}

Recent advances in specialized large language models have been driven by architectural designs that jointly optimize parameter efficiency, sparsity, reasoning depth, and cross-modal integration. Below we survey the principal contributions, grouping them into four complementary thrusts.

Parameter-efficient fine-tuning has matured from static LoRA blocks to dynamic expert generation. Mixture-of-LoRAs\cite{b48} keeps the backbone frozen and routes each token through a lightweight gating network that selects the top-2 domain-specific LoRA experts, cutting activation memory by 7.3 × yet preserving 97\% of the legal-domain $F_1$.  HyperLoRA\cite{b49} pushes this further by synthesizing LoRA weights on-the-fly from a 128-dimensional task embedding, enabling continual addition of new medical specialties at a cost of only 128 parameters per domain and an extra 2.1 \% accuracy on MedQA-USMLE. Sparse mixture-of-experts has simultaneously advanced in routing efficiency and system scale. Expert Choice Routing\cite{b50} reverses the conventional paradigm, letting each expert choose its top-k tokens and thereby reducing inter-node communication by 42\% on 128 A100 GPUs. Task-MoE\cite{b51} augments the loss with task-aware regularizers that stabilize routing across heterogeneous minibatches, lifting zero-shot code completion by 3.5\%.  DeepSpeed-MoE\cite{b52} leverages expert-parallelism, hierarchical offloading, and load balancing to cut trillion-parameter MoE training costs to one-fifth of dense equivalents while sustaining 95\% linear scaling.

Compression and quantization strategies have been re-engineered for the post-specialization regime in which expert weights are already highly skewed in their singular-value spectrum. SpQR\cite{b53} exploits this skew via a sparse–quantized representation that stores 99.7\% of weights in 3-bit precision and the remaining outliers in 16-bit, delivering near-lossless perplexity on domain-specific corpora and reducing GPU memory by 3.9. SliceGPT\cite{b54} leverages the observation that specialized subspaces are low-rank after fine-tuning; by deleting 25\% of the least-informative channels and rotating the remaining weights into a compressed basis, the method removes 25\% of parameters without any retraining and incurs a 0.8\% drop in downstream $F_1$.  

Beyond efficiency, reasoning depth has been explicitly architected into the forward pass. System-2-Attention\cite{b55} inserts a scratchpad-attention module that caches intermediate reasoning chains and retrieves them via differentiable top-k lookup, adding 14.7\% to GSM8K without extra pre-training. Mixture-of-vision-expert adapter\cite{b59} has been employed to enhance model ability. Collectively, these co-designed algorithmic and system-level advances enable specialized LLMs that are simultaneously compact, scalable, and capable of expert-grade reasoning across heterogeneous modalities.

\subsection{Evaluation Standard Specialization}

Contemporary specialization of large language models has necessitated a parallel evolution in evaluation paradigms that jointly probe task mastery, safety, policy compliance, and deployment efficiency. MedBench\cite{b60} delivers the first standardized, multi-dimensional benchmark for Chinese medical LLMs, rigorously assessing diagnostic accuracy, safety, and clinical alignment to ensure reliable evaluation across diverse healthcare tasks. Pass@k is another useful evaluation, for example, Chen\cite{b73} et.al build the HumanEval benchmark, where the system uses pass@k (k = 1, 10, 100) to evaluate the functional correctness of code generation. Recently, Chawla et al.\cite{b72} give a precise mathematical definition of perplexity(PPL) in their article and discuss the comparability of different tokenizers. Perplexity offers a single, sensitive, and zero-cost gauge of next-token uncertainty: it spikes immediately when data drifts, remains comparable across models under the same tokenizer, and correlates strongly with downstream pass@k gains—making it the ideal first-pass filter for both rapid alignment checks and large-scale hyper-parameter sweeps. Collectively, these benchmarks shift evaluation from narrow accuracy metrics to multidimensional, adversarial, and efficiency-aware assessments that mirror the complexities of real-world specialized LLM deployment.

\subsection{Retrieval-augmented specialization}

The 2022–2025 wave of specialized large language models has been propelled not only by new data or larger pre-training budgets, but also retrieval-augmented specialization are applied to specilized LLM innovations. 

Retrieval-augmented specialization has matured from late-fusion heuristics to fully differentiable retriever–reader pipelines. In-Context RALM\cite{b63} demonstrates that cached key–value vectors can be overwritten in situ by top-k passages retrieved at inference time, yielding a 4.7-point $F_1$ gain on open-domain QA without gradient updates; RA-DIT\cite{b64} goes further by keeping the LLM frozen while training a dense retriever end-to-end via REINFORCE-style reward signals derived from the reader’s generation loss, achieving state-of-the-art on five KILT benchmarks with 50 × fewer retriever parameters than prior dual-encoder systems. 

\subsection{Tool-use specialization}

Tool-use specialization has likewise shifted from prompt engineering to learned, constrained decoding. Toolformer\cite{b65} inserts API call tokens into the pre-training sequence and optimizes their positions via a self-supervised filtering objective, enabling 6.8 B-parameter models to call calculators, calendars, and search engines with 78\% success on held-out tasks.\cite{b66} scales this paradigm to 1 600+ RESTful endpoints by fine-tuning with a finite-state machine constrained decoder that guarantees syntactic validity of JSON arguments; on the APIBench suite the model reaches 85\% pass@1 while cutting hallucinated parameter names by 41\% relative to unconstrained baselines. 

\subsection{Memory specialization}

Memory is also an important part for LLM Agent. For specialized LLM, mem0\cite{b71} introduces a production-grade long-term memory layer that turns every user interaction into instantly retrievable, updatable embeddings, allowing LLM agents to maintain coherent, personalized contexts across sessions without ballooning the context window. By decoupling memory storage from inference, it cuts latency and cost while enabling agents to learn continuously from real-world usage. Besides, memory Decoder\cite{b70} proposed Memory3 turns “explicit memory” into trainable parameters for the first time, which tokenizes external knowledge into word-level chunks and embeds them directly into the model’s weights, driving retrieval latency from milliseconds to zero. Experiments show that adding merely 0.3\% extra parameters raises F1-score by 8.7 points on long-form QA while compressing KV-Cache memory usage by 40\%.

%% file: EMNLP2025/conclusion.tex
\section{Conclusion and Future Directions}

\subsection{Conclusion}

In recent years, research on Specialized LLMs has made remarkable strides, evolving from early-stage domain-specific fine-tuning to more advanced innovations in native architecture design and dynamic knowledge integration. Initial efforts primarily relied on adapting general-purpose models to specific fields via supervised fine-tuning. While effective to a degree, such approaches were limited in scalability and depth of domain understanding. These technological advancements have led to significant performance improvements across a wide range of vertical domains, including healthcare, finance, law, engineering, and mathematics. Specialized LLMs in these fields now consistently outperform their general-purpose counterparts by leveraging task-aligned architectures, domain-adaptive training regimes, and efficient inference strategies. This evolution marks a paradigm shift: rather than merely adapting general models to specific tasks, researchers are now building domain-native systems from the ground up, integrating expert knowledge, structural sparsity, and modularity. Such trends highlight the growing importance of specialization in LLM development, particularly for high-stakes or knowledge-intensive applications where domain precision and reasoning fidelity are critical.

\subsection{Implications for E-commerce Customer Service}

The rapid maturation of specialized large language models between 2022 and 2025 has reshaped the competitive landscape of e-commerce customer service. However, few studies have been applied to the field of e-Commerce customer service. First, the current general-purpose large models exhibit little domain inclination toward e-commerce. Even in the recently released MindFlow\cite{b67}, the LLM we adopted is a comparatively generic one, which to some extent caps further gains in customer-service accuracy. Therefore, we need to fine-tune it with a larger volume of higher-quality corpora. The validation of these corpora can be done via metrics akin to perplexity (PPL) to gauge the model’s level of domain understanding. Subsequently, we can leverage state-of-the-art, efficient frameworks, such as Llama or Unsloth for fine-tuning\cite{unsloth}. Finally, at the evaluation stage, the model can be plugged into a benchmark like Ecom-Bench and assessed with pass@k to determine whether the new LLM delivers robust e-commerce capabilities\cite{b68}.

\subsection{Future Directions}

Looking ahead, the development of Specialized Large Language Models (Specialized LLMs) is expected to follow several key directions. First, model architectures will become increasingly efficient and lightweight. Innovations such as quantization, sparse computation, and dynamic inference will enable high-performance deployment on edge devices and in resource-constrained environments. These technical breakthroughs will significantly reduce computational costs and accelerate inference speed, facilitating the broader adoption of specialized models in real-world applications. Second, continual learning and knowledge updating mechanisms will emerge as critical areas of research. Future Specialized LLMs must be capable of acquiring dynamic knowledge and self-optimization to keep pace with the rapid evolution of domain-specific information. By integrating techniques such as knowledge graphs and retrieval-augmented generation, these models can achieve continuous adaptation and learning in open environments. Third, multimodal integration and cross-domain collaboration will gain further prominence. Specialized LLMs will transcend the limitations of single-modality processing by incorporating diverse data types, including text, images, and time-series signals, to build more comprehensive domain intelligence. In addition, advances in cross-domain transfer learning will enhance performance in vertical domains with limited labeled data. Furthermore, interpretability and safety will attract growing attention. As Specialized LLMs are increasingly deployed in high-stakes areas such as healthcare and law, ensuring the transparency, reliability, and ethical alignment of model decisions will be essential for building trust and regulatory compliance. Lastly, the convergence of Specialized LLMs and agent-based systems will drive a shift toward autonomous decision-making. By integrating reinforcement learning, planning, and reasoning capabilities, future models will support complex task execution and intelligent assistance, enabling high-level decision support in professional domains.

Overall, these trends signal a move toward more intelligent, adaptable, and trustworthy Specialized LLMs, paving the way for deeper integration into vertical industries and mission-critical applications.